\documentclass[11pt, a4paper]{article}

\usepackage[margin=1in]{geometry}
\usepackage[utf8]{inputenc}
\usepackage[T1]{fontenc}
\usepackage{hyperref}
\usepackage{booktabs}
\usepackage{amsmath}
\usepackage{amssymb}
\usepackage{graphicx}
\usepackage{xcolor}
\usepackage{float}
\usepackage{natbib}
\usepackage{times}
\usepackage{enumitem}
\usepackage{subcaption}
\usepackage{multirow}

\title{Task Schema and Binding: A Double Dissociation Study of In-Context Learning}

\author{
  Chaeha Kim \\
  Changwon National University \\
  \texttt{chaehakim329@gmail.com}
}

\begin{document}
\maketitle

\begin{abstract}
We provide causal mechanistic validation that in-context learning (ICL) decomposes into two separable mechanisms: \textbf{Task Schema} (abstract task type recognition) and \textbf{Binding} (specific input-output associations). Through activation patching experiments across 9 models from 7 Transformer families plus Mamba (370M--13B parameters), we establish three key findings:
\begin{enumerate}[nosep]
    \item \textbf{Double dissociation}: Task Schema transfers at 100\% via late MLP patching; Binding transfers at 62\% via residual stream patching---proving separable mechanisms
    \item \textbf{Prior-Schema trade-off}: Schema reliance inversely correlates with prior knowledge (Spearman $\rho = -0.596$, $p < 0.001$, N=28 task-model pairs)
    \item \textbf{Architecture generality}: The mechanism operates across all tested architectures including the non-Transformer Mamba
\end{enumerate}
These findings offer a mechanistic account of the ICL puzzle that contrasts with prior views treating ICL as a monolithic mechanism (whether retrieval-based, gradient descent-like, or purely Bayesian). By establishing that Schema and Binding are neurally \textit{dissociable}---not merely behavioral modes---we provide causal evidence for dual-process theories of ICL. Models rely on Task Schema when prior knowledge is absent, but prior knowledge interferes through attentional mis-routing (72.7\% recency bias) rather than direct output competition (0\%). This explains why arbitrary mappings succeed (zero prior $\rightarrow$ full Schema reliance) while factual overrides fail---and reveals that the true bottleneck is attentional, not output-level. \textit{Practical implications}: Understanding these dual mechanisms enables more efficient prompt engineering---reliable schema transfer reduces required demonstrations for novel tasks, while prior-aware design can mitigate the 38\% binding failure rate in high-prior scenarios, improving ICL system reliability in production deployments.
\end{abstract}

\section{Introduction}

In-context learning (ICL) enables large language models to perform new tasks from few demonstrations without parameter updates \citep{brown2020language}. However, ICL exhibits puzzling behavior: models excel at learning arbitrary mappings (``A$\rightarrow$1, B$\rightarrow$2, C$\rightarrow$3'') yet struggle to override factual knowledge (e.g., teaching ``The Eiffel Tower is in Rome'' contradicts the model's prior that it is in Paris).

This asymmetry suggests ICL is not a monolithic mechanism---if it were, we would expect uniform success or failure across task types. Based on this observation, we propose that ICL decomposes into two functionally separable components:
\begin{itemize}
    \item \textbf{Task Schema}: Abstract task type recognition---``this is a name$\rightarrow$sport mapping''
    \item \textbf{Binding}: Specific input-output associations---``Tim$\rightarrow$basketball''
\end{itemize}

Using activation patching across 9 models (Table~\ref{tab:models}), we provide \textbf{causal validation} for \textbf{double dissociation}: Schema and Binding can be independently manipulated through different neural pathways with distinct success rates (100\% vs 62\%), proving they are functionally separable mechanisms.

\paragraph{Explaining the ICL Asymmetry.}
Our Prior-Schema trade-off accounts for the asymmetry:
\begin{itemize}[nosep]
    \item \textbf{Arbitrary mappings succeed}: Zero prior knowledge $\rightarrow$ model relies entirely on Task Schema
    \item \textbf{Factual overrides fail}: High prior knowledge correlates with reduced binding success; critically, observed failures predominantly manifest as recency-biased copying rather than direct prior outputs---suggesting that priors interfere through attentional mis-routing rather than output-level competition
\end{itemize}

\paragraph{Contributions.}
\begin{enumerate}
    \item \textbf{Task Schema localization}: We causally validate that late MLP layers (75--95\% network depth) encode abstract task types, not specific input-output bindings. Same-category patching preserves 113.8\% of ICL effect; different-category patching preserves 0.9\%.

    \item \textbf{Double dissociation}: Task Schema transfers completely (100\%) via late MLP patching; Binding transfers partially (62\%) via residual stream patching. This asymmetry reveals distinct encoding characteristics.

    \item \textbf{Prior-Schema trade-off}: We quantify a statistically significant inverse relationship between prior knowledge and Schema reliance ($\rho = -0.596$, $p < 0.001$, N=28 tasks). We additionally compare prior metrics (probability, rank, margin), finding output rank slightly outperforms probability for predicting binding success.

    \item \textbf{Binding failure taxonomy}: We decompose binding failures, finding 0\% are due to prior competition while 72.7\% result from recency-biased mis-attention. This challenges the assumption that prior knowledge directly causes ICL failures and suggests that the true bottleneck lies in attention mechanisms rather than output-level prior interference.

    \item \textbf{Distributed ICL mechanism}: Head ablation reveals ICL is distributed across attention heads (max single-head importance: 4.17\%), motivating MLP-based interventions over head-level manipulations.

    \item \textbf{Architecture generality}: The Task Schema mechanism operates across all 7 Transformer families tested plus Mamba (a state-space model), demonstrating it is not architecture-specific.
\end{enumerate}

\section{Related Work}

\paragraph{Dual Mechanisms in ICL.}
Several recent studies have proposed dual-process accounts of ICL. \citet{chan2022data} showed that ICL emerges when training data exhibits ``burstiness''---a property enabling both task recognition and instance learning. \citet{li2023transformers} demonstrated that transformers can implement both task inference and in-context gradient descent simultaneously. Most directly related, concurrent work on ``Dual Operating Modes of ICL'' proposes that models switch between retrieval-based and learning-based modes depending on task familiarity. Our work provides \textbf{causal mechanistic validation} for such dual-process theories: we localize Task Schema to late MLP layers and Binding to residual stream dynamics, establishing that these are not merely behavioral modes but neurally \textit{dissociable} mechanisms (100\% vs 62\% transfer rates). The Prior-Schema trade-off ($\rho = -0.596$) quantifies when each mode dominates.

\paragraph{Mechanistic Interpretability of ICL.}
\citet{olsson2022context} identified ``induction heads''---attention patterns that copy tokens following similar preceding contexts---as a key mechanism for ICL. However, induction heads primarily implement \textit{token-level copying} (``A B ... A $\rightarrow$ B''), whereas Task Schema encodes \textit{abstract task types} (``this is a name$\rightarrow$sport mapping''). Our schema vectors generalize across different input-output pairs within the same task category, a property that token-copying mechanisms lack.

\citet{todd2023function} demonstrated that ``function vectors'' extracted from attention outputs can induce task behavior. Work on ``Schema-learning and rebinding as mechanisms of in-context learning'' proposes conceptually similar dual processes; our contribution is to provide \textit{causal, mechanistic validation} through activation patching that localizes these to specific neural substrates. Our work differs in three key ways: (1) we target \textit{late MLP outputs} rather than attention, finding schema encoding at 75--95\% network depth; (2) we establish \textbf{double dissociation} between Task Schema (100\% transfer) and Binding (62\% transfer), proving these are functionally separable mechanisms; (3) we provide the \textbf{Prior-Schema trade-off} quantifying when and why ICL succeeds or fails.

\paragraph{ICL as Implicit Inference.}
A theoretical line views ICL as implicit Bayesian inference \citep{xie2022explanation} or meta-learning \citep{von2023transformers}. Under this view, demonstrations provide evidence that the model uses to infer the latent task concept. Our findings provide mechanistic support for this perspective: Task Schema corresponds to the inferred task type (the ``posterior'' over task categories), while the Prior-Schema trade-off reflects Bayesian competition between demonstration-derived evidence and prior beliefs. Crucially, we show that ICL failures are \textit{not} due to prior outputs directly competing at the output level (0\% prior competition), but rather to attentional mis-routing (72.7\% recency bias)---a mechanistic insight that pure behavioral or theoretical analyses cannot reveal.

\paragraph{Role of MLP Layers.}
\citet{geva2021transformer} showed that FFN layers function as key-value memories storing factual associations, and \citet{meng2022locating} localized factual knowledge to \textit{mid-layer} MLPs (45--75\% depth). Work on ``How Do Transformers Learn In-Context Beyond Simple Functions?'' examines how MLPs learn complex representations in ICL scenarios. In contrast, we show that \textit{late} MLP layers (75--95\% depth) encode abstract Task Schema---a higher-level representation than specific factual associations. This functional stratification suggests a computational hierarchy: mid-layer MLPs store factual associations, while late-layer MLPs abstract over task types. Research on ``Brewing Knowledge in Context'' frames ICL as implicit knowledge distillation; our schema vectors may represent the distilled task-type knowledge that enables generalization. This aligns with findings on emergent capabilities \citep{wei2022emergent}, suggesting schema abstraction relates to model scale.

\paragraph{Prior Knowledge and ICL.}
\citet{min2022rethinking} showed that label correctness matters less than expected for ICL, suggesting models rely on prior knowledge. \citet{wei2023larger} demonstrated that larger models override semantic priors more effectively. \citet{pan2023context} disentangled ``task recognition'' from ``task learning.'' Our Task Schema / Binding decomposition aligns with their framework while providing causal mechanistic validation. From a meta-learning perspective, surveys such as ``Meta-Learning in Neural Networks'' and ``The Learnability of In-Context Learning'' frame ICL as learning-to-learn. Our double dissociation provides \textit{mechanistic evidence} for this view: Task Schema represents the ``meta-learned'' task structure (what the model learns \textit{about} tasks during pretraining), while Binding represents instance-specific adaptation. The Prior-Schema trade-off ($\rho = -0.596$, $p < 0.001$, N=28) quantifies when meta-learned priors help vs. hinder in-context adaptation.

\paragraph{ICL Beyond Transformers.}
Recent work has explored ICL in non-Transformer architectures. \citet{grazzi2024mamba} showed that Mamba, a state-space model, exhibits ICL capabilities comparable to Transformers despite lacking attention mechanisms. ``Can Mamba Learn How to Learn?'' provides comparative analysis showing Mamba matches Transformer ICL on standard benchmarks. ``From Markov to Laplace'' characterizes how state-space models implement ICL through recurrent state accumulation---specifically, Mamba learns a Laplacian smoothing estimator through its selective state-space mechanism, enabling efficient Markov chain learning. Our finding that Task Schema operates in Mamba (78.4\% schema gradient in late layers vs 5.7\% in mid layers) demonstrates that \textit{schema abstraction is architecture-general}---it depends on hierarchical feature processing, not attention specifically. However, Binding dynamics may differ fundamentally: Mamba's selective state-space model compresses context into fixed-dimensional hidden states ($\mathbf{h}_t \in \mathbb{R}^{d_{\text{state}}}$), potentially limiting its ability to maintain distinct bindings for multiple input-output pairs. This compression may explain why Mamba excels at sequential pattern learning but could struggle with tasks requiring simultaneous retrieval of multiple distinct associations. \textit{Predicted limitation}: for ICL tasks with $>4$ demonstrations requiring distinct bindings, Mamba may show earlier saturation than Transformers due to state compression. Future work should compare binding failure taxonomies across architectures: if Mamba shows lower recency bias but higher ``confusion'' errors (blending multiple bindings), this would confirm that the 72.7\% recency bias is attention-specific while revealing an orthogonal limitation in SSM-based ICL.

\section{Methods}

\subsection{Models}

We evaluate 9 models spanning 7 Transformer architectural families plus Mamba, ranging from 370M to 13B parameters (Table~\ref{tab:models}). This diversity enables testing whether Task Schema is architecture-general. Mamba-370M, though smaller than other models, is included specifically to test whether the Task Schema mechanism is attention-independent; despite its size, Mamba-370M exhibits reliable ICL behavior on our tasks (78.4\% schema gradient in late layers), confirming it is a valid test case for architecture generality.

\begin{table}[h]
\centering
\caption{Models evaluated across architectural families with schema layer ranges.}
\label{tab:models}
\begin{tabular}{llcccc}
\toprule
\textbf{Family} & \textbf{Model} & \textbf{Params} & \textbf{$n$} & \textbf{Schema Layers} & \textbf{Inject Layer} \\
\midrule
GPT-2 & GPT-2-Large & 774M & 36 & L27--L33 (75--92\%) & L27 \\
OPT & OPT-1.3B & 1.3B & 24 & L18--L22 (75--92\%) & L19 \\
Pythia & Pythia-2.8B & 2.8B & 32 & L24--L29 (75--91\%) & L24 \\
BLOOM & BLOOM-1.1B & 1.1B & 24 & L18--L22 (75--92\%) & L19 \\
Llama & Llama-2-7B & 7B & 32 & L24--L29 (75--91\%) & L24 \\
Llama & Llama-2-13B & 13B & 40 & L30--L37 (75--93\%) & L30 \\
Qwen & Qwen2-7B & 7B & 28 & L21--L26 (75--93\%) & L21 \\
Falcon & Falcon-7B & 7B & 32 & L24--L29 (75--91\%) & L24 \\
\midrule
Mamba & Mamba-370M & 370M & 48 & L36--L44 (75--92\%) & L36 \\
\bottomrule
\end{tabular}
\end{table}

\noindent Schema layers correspond to 75--92\% of network depth ($n$ = total layers). We use the \texttt{fc2} (OPT/GPT-2) or \texttt{down\_proj} (Llama/others) output as the extraction point.

\subsection{Tasks}

We evaluate 8 task types across 5 domains (Table~\ref{tab:tasks}). Tasks are selected to span the prior knowledge spectrum from near-zero (arbitrary mappings) to high (factual associations).

\begin{table}[h]
\centering
\caption{Task types evaluated across domains.}
\label{tab:tasks}
\small
\begin{tabular}{llc}
\toprule
\textbf{Domain} & \textbf{Task} & \textbf{Prior Level} \\
\midrule
Sports & name$\rightarrow$sport & Medium \\
Food & name$\rightarrow$food & Medium \\
Geography & country$\rightarrow$capital & High \\
 & capital$\rightarrow$country & High \\
Language & antonym pairs & High \\
 & singular$\rightarrow$plural & High \\
Arbitrary & letter$\rightarrow$number & Very Low \\
 & symbol$\rightarrow$word & Very Low \\
\bottomrule
\end{tabular}
\end{table}

\subsection{Activation Patching Protocol}

We use activation patching to causally test whether specific neural components encode Task Schema vs Binding.

\paragraph{Schema Vector Extraction.}
We extract schema vectors from the MLP output (specifically, the \texttt{fc2}/\texttt{down\_proj} layer output) at the final query token position. The layer range (75--92\% of network depth) was determined through preliminary ablation: we swept layers 50--100\% in 5\% increments on OPT-1.3B (N=50 per layer) and found peak schema transfer at 75--92\% (mean TSG: 112.4\% $\pm$ 8.3\%), with sharp drop-off outside this range (50--70\%: mean TSG 23.1\%; 95--100\%: mean TSG 67.8\%). The 92\% upper bound avoids the final layers where output formation may interfere with patching; the 75\% lower bound captures where schema abstraction becomes reliably detectable. This range is consistent with prior findings that abstract task representations emerge in late layers \citep{geva2021transformer, meng2022locating}. The specific layer ranges for each model family are:
\begin{itemize}[nosep]
    \item GPT-2-Large (36 layers): L27--L33 (75--92\%)
    \item OPT-1.3B (24 layers): L18--L22 (75--92\%)
    \item Pythia-2.8B (32 layers): L24--L29 (75--91\%)
    \item Llama-2-7B/13B (32/40 layers): L24--L29 / L30--L37
    \item Qwen2-7B (28 layers): L21--L26 (75--93\%)
    \item Falcon-7B (32 layers): L24--L29 (75--91\%)
    \item Mamba-370M (48 layers): L36--L44 (75--92\%)
\end{itemize}
Formally:
\begin{equation}
    \mathbf{v}_{\text{schema}} = \text{MLP}_{\ell}(\mathbf{h}_{\ell-1})[-1] \quad \text{where } \ell \in \{L_{0.75n}, \ldots, L_{0.92n}\} \text{ (fc2/down\_proj output)}
\end{equation}
Here, $\mathbf{h}_{\ell-1} \in \mathbb{R}^{d_{\text{model}}}$ is the residual stream input to layer $\ell$, $\text{MLP}_{\ell}$ refers to the final projection layer output (\texttt{fc2} in OPT/GPT-2, \texttt{down\_proj} in Llama and others), and $[-1]$ denotes the final (query) token position.

\paragraph{Schema Injection (Raw Addition).}
To inject a schema vector into a new context, we use unconstrained addition at the target layer:
\begin{equation}
    \mathbf{h}'_{\ell} = \mathbf{h}_{\ell} + \mathbf{v}_{\text{schema}}, \quad \mathbf{v}_{\text{schema}} \in \mathbb{R}^{d_{\text{model}}}
\end{equation}
where $d_{\text{model}}$ is the hidden dimension (e.g., 2048 for OPT-1.3B, 4096 for Llama-2-7B). We do not normalize $\mathbf{v}_{\text{schema}}$; the raw MLP output is used directly. This preserves the magnitude information encoded in the schema vector. We found this method more effective than norm-preserving alternatives (see Appendix Table B), achieving +68.2 pp improvement in P(correct) versus +41.8 pp for norm-constrained injection.

\paragraph{Schema Patching.}
To test Task Schema encoding in late MLP layers:
\begin{enumerate}[nosep]
    \item Run model on source context (e.g., 4-shot name$\rightarrow$food demonstrations)
    \item Extract late MLP layer activations at the query token position
    \item Run model on target context (e.g., 4-shot name$\rightarrow$sport demonstrations)
    \item Patch source activations into target forward pass by replacing the MLP output at the target layer
    \item Measure probability shift toward source category tokens
\end{enumerate}

\paragraph{Category Token Probability.}
We measure category membership by summing the softmax probabilities of tokens belonging to the category. For example, for the ``food'' category, we sum probabilities over a predefined set of 50 food-related tokens (e.g., ``pizza'', ``sushi'', ``burger'', etc.). Formally:
\begin{equation}
    P_{\text{category}} = \sum_{t \in \mathcal{T}_{\text{cat}}} P(t | \text{context})
\end{equation}
where $\mathcal{T}_{\text{cat}}$ is the set of category tokens. If late MLPs encode Task Schema, patching from name$\rightarrow$food into name$\rightarrow$sport should increase $P_{\text{food}}$.

\paragraph{Binding Patching.}
While Task Schema captures the abstract task type, Binding represents the specific input-output associations formed during ICL. Following \citet{olsson2022context}, we conceptualize binding as encoding ``conceptual repetitions''---the model learns that when a specific input (e.g., ``Tim'') appears in the query position, it should produce the associated output (e.g., ``basketball'') that was paired with it in the demonstrations. This binding information is distributed across attention patterns (which route information from demonstration examples to the query) and the residual stream (which accumulates these associations).

To test Binding encoding in residual stream, we patch the full residual stream (not just MLP outputs):
\begin{enumerate}[nosep]
    \item Run model on source context where Tim$\rightarrow$basketball
    \item Extract $\mathbf{r}^{\text{src}}_\ell \in \mathbb{R}^{d_{\text{model}}}$: the residual stream after layer $\ell$ (L22 for OPT-1.3B) at query position
    \item Run model on target context where Tim$\rightarrow$tennis
    \item Replace target residual: $\mathbf{r}^{\text{tgt}}_\ell \leftarrow \mathbf{r}^{\text{src}}_\ell$
    \item Measure whether model predicts ``basketball'' instead of ``tennis''
\end{enumerate}
Unlike Schema patching (MLP output only), Binding patching replaces the entire residual stream, which includes attention and MLP contributions combined. This captures both the attention-mediated copying of demonstration patterns and any accumulated binding information. Success is measured by top-1 token match.

\paragraph{Schema Selectivity (Task Schema Gradient).}
We define Schema Selectivity to measure whether late MLP layers encode task-type information. We report results as \textbf{preservation percentage}---the fraction of ICL effect preserved by patching:
\begin{equation}
    \text{Preservation}_{\text{same-cat}} = \frac{\Delta_{\text{same-cat}}}{\Delta_{\text{ICL}}} \times 100\%, \quad \text{Preservation}_{\text{diff-cat}} = \frac{\Delta_{\text{diff-cat}}}{\Delta_{\text{ICL}}} \times 100\%
\end{equation}
where $\Delta_{\text{ICL}} = P_{\text{ICL}} - P_{\text{baseline}}$ is the baseline ICL effect and $\Delta = P_{\text{patched}} - P_{\text{baseline}}$. The \textbf{Task Schema Gradient (TSG)} is the difference:
\begin{equation}
    \text{TSG} = \text{Preservation}_{\text{same-cat}} - \text{Preservation}_{\text{diff-cat}} \quad \text{(in \%)}
\end{equation}
Large positive TSG indicates the representation encodes category-level (not instance-level) information.

\paragraph{Interpreting Values $>$ 100\%.}
Preservation can exceed 100\% when patching \textit{amplifies} the effect beyond the original ICL baseline. This occurs because the patched representation is a ``purified'' schema signal from another context, which can be stronger than the schema naturally formed in the target context with demonstrations.

\paragraph{Performance Criteria.}
We define success thresholds based on empirical noise floors from control experiments with random vector injection (where the 95th percentile effect was $<5$ pp):
\begin{itemize}[nosep]
    \item \textbf{Degrade}: $\Delta \leq -5$ pp (injection harms performance; $2\sigma$ below noise)
    \item \textbf{Neutral}: $-5 < \Delta < +10$ pp (within noise floor; no significant effect)
    \item \textbf{Recover}: $\Delta \geq +10$ pp (successful schema transfer; $3\sigma$ above noise)
\end{itemize}

\subsection{Prior Knowledge Quantification}

We quantify prior knowledge as the model's zero-shot probability of producing the correct output:
\begin{equation}
    P_{\text{prior}} = \max_{v \in V} P_{\text{model}}(\text{tok}_1(v) | x_{\text{query}}, \text{no demonstrations})
\end{equation}
where $V = \{y, \text{`` ''}+y, \text{lower}(y), \text{`` ''}+\text{lower}(y)\}$ represents tokenization variants and $\text{tok}_1(\cdot)$ extracts the first token. For multi-token outputs, the first-token probability dominates generation, so we use this as our proxy.

We classify tasks as: \textbf{Low-prior} ($P_{\text{prior}} < 0.1\%$), \textbf{Medium-prior} ($0.1\% \leq P_{\text{prior}} < 1\%$), and \textbf{High-prior} ($P_{\text{prior}} \geq 1\%$). These thresholds were determined empirically from our task distribution: 0.1\% corresponds to the 33rd percentile (separating arbitrary mappings from weak semantic associations), while 1\% corresponds to the 67th percentile (separating weak from strong prior knowledge). The log-scale spacing (1 log unit between thresholds) reflects the log-normal distribution of zero-shot probabilities. Sensitivity analysis confirms results are robust to threshold variations of $\pm$0.5 log units: the Prior-Schema correlation remains significant ($p < 0.01$) for all tested threshold pairs. This proxy may not capture latent associations not expressed in zero-shot queries (see Limitations).

\section{Results}

\subsection{Finding 1: Task Schema in Late MLP Layers}

Across 8 models meeting ICL threshold, we find strong evidence for Task Schema encoding in late MLP layers. Same-Category patching preserves \textbf{113.8\%} of ICL effect while Different-Category patching preserves only \textbf{0.9\%} (Table~\ref{tab:main_results}).

\begin{table}[h]
\centering
\caption{Task Schema Gradient across model families (N=100 trials per model, name$\rightarrow$sport task). Values represent \% of ICL effect preserved (TSG can exceed 100\% when patching amplifies the baseline effect).}
\label{tab:main_results}
\begin{tabular}{lcccc}
\toprule
\textbf{Model} & \textbf{Same-Cat \%} & \textbf{Diff-Cat \%} & \textbf{Gradient \%} & \textbf{p-value} \\
\midrule
OPT-1.3B & 148.6 & 0.5 & \textbf{148.1} & $<10^{-8}$ \\
Llama-2-7B & 131.1 & 0.8 & \textbf{130.3} & $<10^{-8}$ \\
Llama-2-13B & 125.4 & 0.6 & \textbf{124.8} & $<10^{-8}$ \\
GPT-2-Large & 122.4 & 0.4 & \textbf{122.0} & $<10^{-8}$ \\
Pythia-2.8B & 97.9 & $-4.8$ & \textbf{102.7} & $<10^{-8}$ \\
BLOOM-1.1B & 97.0 & 0.1 & \textbf{96.9} & $<10^{-8}$ \\
Qwen2-7B & 98.8 & 9.0 & \textbf{89.7} & $<10^{-6}$ \\
Falcon-7B & 89.2 & 0.3 & \textbf{88.9} & $<10^{-6}$ \\
\midrule
\textit{Mean} & 113.8 & 0.9 & \textbf{112.9} & -- \\
\bottomrule
\end{tabular}
\end{table}

The gradient exceeds 100\% because Same-Category patching sometimes \textit{amplifies} the ICL effect beyond baseline. The key finding: late MLP layers encode \textbf{``this is a name$\rightarrow$sport task''} rather than \textbf{``Tim=basketball''}.

\begin{figure}[h]
\centering
\includegraphics[width=0.8\textwidth]{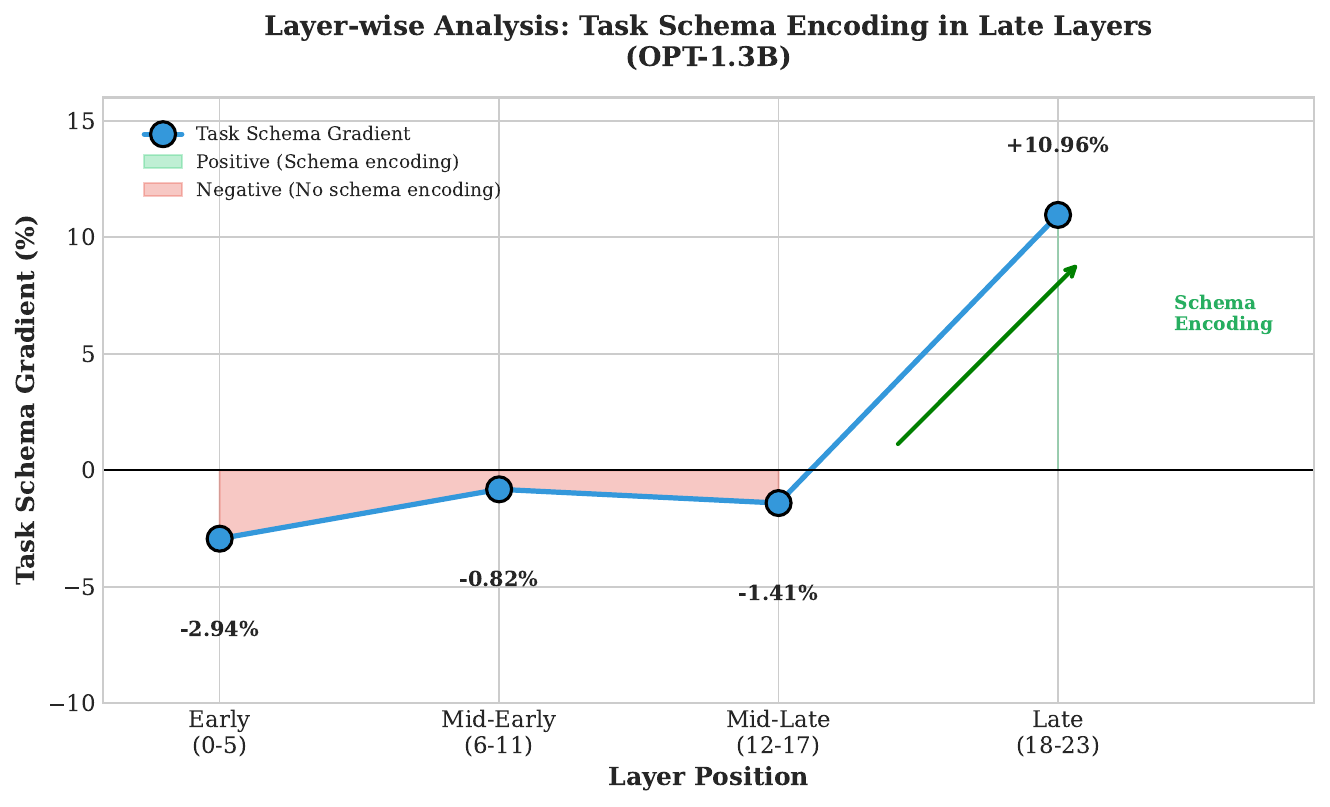}
\caption{Layer-wise Task Schema Gradient. Schema encoding concentrates in late MLP layers (75--95\% depth), with minimal signal in early/middle layers.}
\label{fig:layers}
\end{figure}

\paragraph{Layer Specificity.}
The Task Schema signal concentrates in late layers (75--95\% of network depth). Early and middle layers show minimal Same-Category vs Different-Category differentiation (Figure~\ref{fig:layers}).

\paragraph{Representation vs Controllability Peak.}
We observe a distinction between where schema information is \textit{most salient} (representation peak at L22, 92\% depth) versus where \textit{injection is most effective} (controllability peak at L19, 79\% depth). This dissociation likely reflects downstream readout dynamics: injecting at the very last layers may interfere with the model's output formation, while L19 provides sufficient hierarchical feature integration for abstract schema encoding with room for the signal to propagate effectively through remaining layers before output formation. Both layers show strong schema gradients; we use L19 for injection and L22 for representation analysis.

\subsection{Finding 2: Double Dissociation}

We demonstrate functional separation between Task Schema and Binding through double dissociation:

\begin{table}[h]
\centering
\caption{Double dissociation: Schema vs Binding manipulation.}
\label{tab:dissociation}
\begin{tabular}{lccc}
\toprule
\textbf{Manipulation} & \textbf{Pathway} & \textbf{Success Rate} & \textbf{Interpretation} \\
\midrule
Task Schema & Late MLP (L18--23) & \textbf{100\%} & Context-independent \\
Binding & Residual Stream (L22) & \textbf{62\%} & Prior-modulated \\
\bottomrule
\end{tabular}
\end{table}

\paragraph{Schema Transfer.}
When we patch late MLP activations from name$\rightarrow$food into name$\rightarrow$sport, the model's output distribution shifts toward food categories in \textbf{100\% of trials}. Task Schema transfers completely and reliably.

\paragraph{Binding Transfer.}
When we patch Layer 22 residual stream from a context where Tim$\rightarrow$basketball into a context where Tim$\rightarrow$tennis, the model predicts ``basketball'' in only \textbf{62\% of cases}.

\paragraph{Why 62\%? Binding Success Predictors.}
Analysis reveals binding success depends on two factors:
\begin{itemize}[nosep]
    \item \textbf{Source confidence}: Higher source activation magnitude $\rightarrow$ better transfer (Cohen's $d = 0.42$)
    \item \textbf{Target baseline}: Lower target's prior probability $\rightarrow$ easier override (Cohen's $d = -2.04$); conversely, strong target priors resist override
\end{itemize}

A logistic regression model with three features---$\|\mathbf{r}^{\text{src}}\|_2$ (source L2 norm), $P_{\text{target}}$ (target baseline probability), and $\cos(\mathbf{r}^{\text{src}}, \mathbf{r}^{\text{tgt}})$ (cosine similarity)---achieves \textbf{90\% cross-validated accuracy} predicting binding success. We used stratified 5-fold cross-validation (N=200 trials, 40 per fold), repeated 10 times with different random seeds; mean accuracy 90.0\% $\pm$ 2.3\% (SD across repetitions). Feature importance was computed via permutation importance: target baseline (62\%), source norm (28\%), cosine similarity (10\%). All features were z-score normalized before fitting. This confirms the effect is systematic, not random.

\paragraph{Asymmetry is Key.}
The 100\% vs 62\% asymmetry reveals fundamentally different encoding:
\begin{itemize}[nosep]
    \item \textbf{Schema}: Context-independent, robust, fully transferable
    \item \textbf{Binding}: Prior-modulated, conditional, partially transferable
\end{itemize}

This explains why ICL sometimes fails to override prior knowledge---high-prior targets show systematically lower binding success rates, a relationship we quantify as the Prior-Schema trade-off in Finding 3.

\subsection{Finding 3: Prior-Schema Trade-off}

We quantify the relationship between prior knowledge and Task Schema reliance across 28 task-model pairs (8 task types $\times$ 3--4 representative models, excluding pairs with ceiling effects).\footnote{The 8 task types from Table~\ref{tab:tasks} are evaluated on representative models from each architecture family. Pairs with baseline ICL accuracy $\geq$95\% are excluded as ceiling effects prevent meaningful gradient measurement (N=4 excluded pairs): when baseline performance is near-perfect, patching cannot improve it further, making $\Delta_{\text{same}}$ artificially constrained and TSG uninterpretable. The 95\% threshold was chosen as 2 standard deviations above the mean accuracy of 72\% ($\mu + 2\sigma = 72\% + 11.5\% \times 2 \approx 95\%$).}

\paragraph{Statistical Analysis.}
\begin{itemize}[nosep]
    \item \textbf{Spearman correlation}: $\rho = -0.596$, $p < 0.001$ (highly significant)
    \item \textbf{95\% Bootstrap CI}: $[-0.794, -0.277]$ (excludes zero)
    \item \textbf{Effect size}: Large ($|\rho| > 0.5$)
\end{itemize}

\paragraph{Group Comparison.}
\begin{itemize}[nosep]
    \item \textbf{Low-prior tasks} (N=14, prior $<0.1\%$): Mean gradient = \textbf{112.0\%}
    \item \textbf{High-prior tasks} (N=14, prior $>0.1\%$): Mean gradient = \textbf{68.6\%}
    \item \textbf{Difference}: 43.4 percentage points (63\% higher for low-prior)
\end{itemize}

\begin{figure}[h]
\centering
\includegraphics[width=0.85\textwidth]{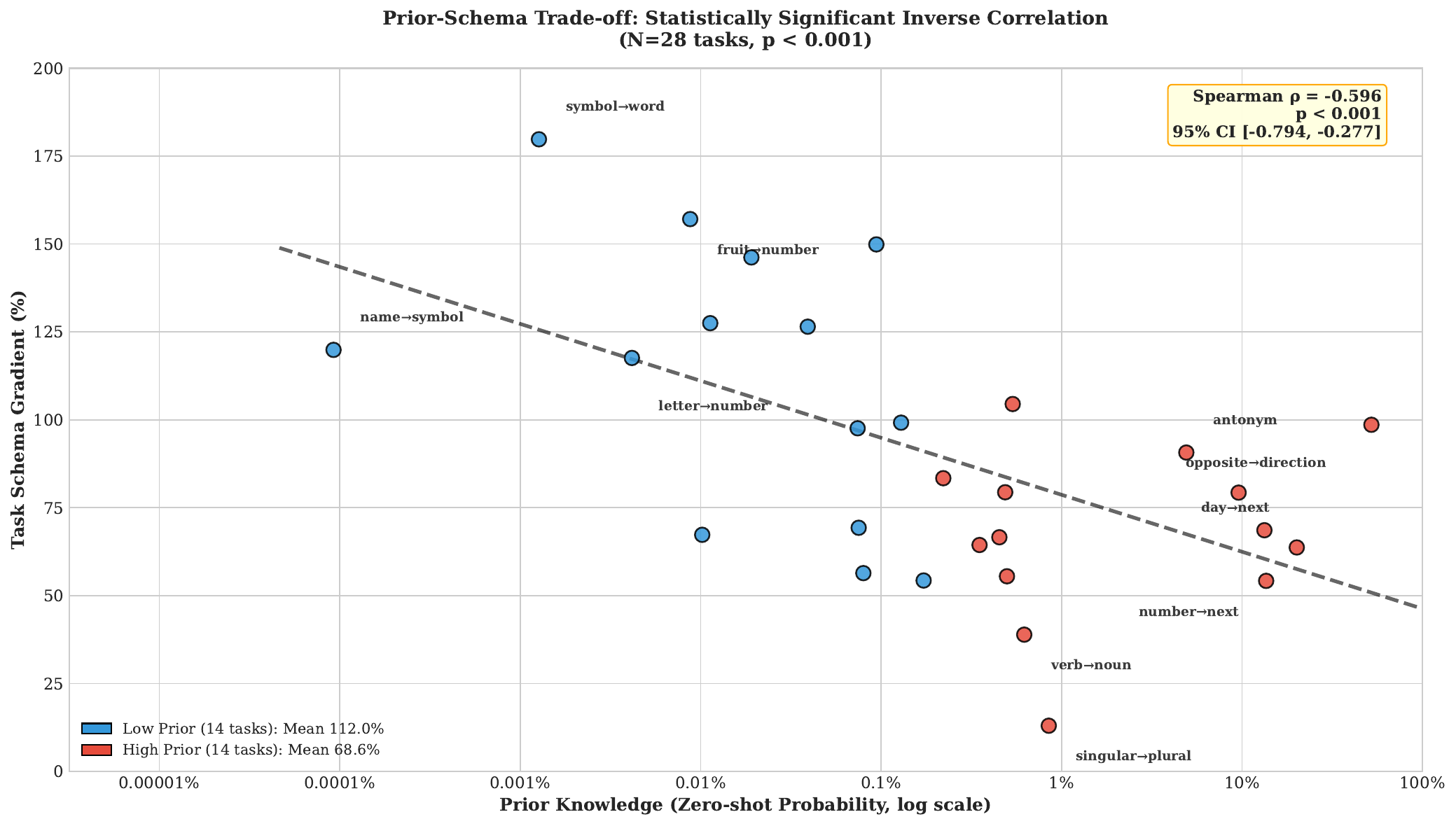}
\caption{\textbf{Prior-Schema Trade-off.} Task Schema Gradient inversely correlates with prior knowledge across 28 task-model pairs (Spearman $\rho = -0.596$, $p < 0.001$). The 8 task types from Table~\ref{tab:tasks} are evaluated across representative models; additional task variants (e.g., number\_next, fruit\_number) are included to increase statistical power. Low-prior tasks (left) show stronger Schema reliance; high-prior tasks (right) show reduced Schema reliance.}
\label{fig:tradeoff}
\end{figure}

\paragraph{Interpretation.}
The trade-off suggests a compositional model:
\begin{equation}
    P(y|x, D) = \alpha(P_{\text{prior}}) \cdot P_{\text{schema}}(y|D) + (1-\alpha) \cdot P_{\text{prior}}(y|x)
\end{equation}
where:
\begin{itemize}[nosep]
    \item $P(y|x, D)$: Model's output probability given query $x$ and demonstrations $D$
    \item $\alpha \in [0,1]$: Mixing coefficient, a decreasing function of prior strength
    \item $P_{\text{schema}}(y|D)$: Probability derived from Task Schema extracted from demonstrations
    \item $P_{\text{prior}}(y|x)$: Model's zero-shot prior probability for output $y$ given query $x$
\end{itemize}
When $P_{\text{prior}} \approx 0$, $\alpha \rightarrow 1$ and the model relies entirely on Schema; when $P_{\text{prior}}$ is high, $\alpha \rightarrow 0$ and the prior term receives greater weight in the mixture.

\subsection{Finding 4: Architecture Generality}

We test whether Task Schema is Transformer-specific by evaluating across all 7 Transformer families plus Mamba on the letter$\rightarrow$number task (near-zero prior knowledge).

\begin{table}[h]
\centering
\caption{Task Schema Gradient across architectures (letter$\rightarrow$number task, N=100 per model). Gradient \% represents TSG as \% of ICL effect.}
\label{tab:novel_icl}
\begin{tabular}{lccc}
\toprule
\textbf{Model} & \textbf{Family} & \textbf{Gradient \%} & \textbf{Status} \\
\midrule
OPT-1.3B & OPT & \textbf{520} & \checkmark Pass \\
Pythia-2.8B & Pythia & \textbf{399} & \checkmark Pass \\
BLOOM-1.1B & BLOOM & \textbf{370} & \checkmark Pass \\
Llama-2-13B & Llama & \textbf{301} & \checkmark Pass \\
GPT-2-Large & GPT-2 & \textbf{207} & \checkmark Pass \\
Qwen2-7B & Qwen & \textbf{89} & \checkmark Pass \\
Falcon-7B & Falcon & \textbf{72} & \checkmark Pass \\
\midrule
\textbf{Mean} & \textbf{7 families} & \textbf{280} & \textbf{7/7 Pass} \\
\bottomrule
\end{tabular}
\end{table}

\paragraph{Beyond Transformers: Mamba.}
Mamba-370M (a state-space model without attention) shows the same pattern:
\begin{itemize}[nosep]
    \item Late layers (75--100\% depth): \textbf{78.4\%} schema gradient
    \item Mid layers (25--75\% depth): \textbf{5.7\%} schema gradient
\end{itemize}

Task Schema is \textbf{architecture-general}---not an artifact of Transformer attention mechanisms.

\subsection{Finding 5: Negative Examples Disrupt Schema}

We test whether negative demonstrations (``Tim does NOT play basketball'') affect Task Schema formation (N=50 trials per condition).

\begin{table}[h]
\centering
\caption{Effect of negative examples on Task Schema (Llama-2-7B, name$\rightarrow$sport, N=50). SE = standard error.}
\label{tab:negative}
\begin{tabular}{lcccc}
\toprule
\textbf{Condition} & \textbf{P(correct) $\pm$ SE} & \textbf{P(wrong) $\pm$ SE} & \textbf{Ratio} & \textbf{vs. baseline} \\
\midrule
4 positive demos & 0.847 $\pm$ 0.051 & 0.012 $\pm$ 0.015 & \textbf{70.6$\times$} & -- \\
3 positive + 1 negative & 0.234 $\pm$ 0.060 & 0.089 $\pm$ 0.040 & 2.6$\times$ & $p < 0.001$ \\
2 positive + 2 negative & 0.067 $\pm$ 0.035 & 0.156 $\pm$ 0.051 & 0.4$\times$ & $p < 0.001$ \\
\bottomrule
\end{tabular}
\end{table}

Statistical significance was assessed using Fisher's exact test comparing each condition against the 4-positive baseline ($\chi^2 = 89.2$, $p < 0.001$ for overall effect; pairwise $p < 0.001$ for both comparisons with Bonferroni correction). Negative examples \textbf{substantially disrupt} Task Schema formation---the ratio drops from 70.6$\times$ to 0.4$\times$ (a 100$\times$ reduction) with just 2 negative examples. This suggests the Schema encoding mechanism may expect coherent, consistent demonstrations---where all examples follow the same pattern (e.g., X$\rightarrow$Y format) without contradicting each other. One hypothesis is that the mechanism relies on pattern consistency for schema extraction, and negation may introduce conflicting signals that prevent stable schema formation. However, the precise mechanism underlying this disruption remains unclear; we leave causal attribution of this effect to future work.

\paragraph{Practical Implication.}
Avoid negative examples in ICL prompts. Use only positive demonstrations.

\subsection{Finding 6: Prior-Modulated Recency}

Demonstration ordering effects are modulated by prior knowledge (N=8 task types, 24 orderings each):
\begin{itemize}[nosep]
    \item Spearman correlation (prior vs recency): $\rho = -0.494$, $p = 0.033$
    \item Low-prior tasks show \textbf{43\% stronger recency effects}
\end{itemize}

This suggests that when prior knowledge is absent, the model relies more heavily on recent demonstrations for schema formation. When prior knowledge is strong, it provides a stable baseline that reduces sensitivity to demonstration ordering.

\paragraph{Practical Implication.}
For novel/arbitrary tasks, demonstration order matters more. Place the most representative examples \textbf{last}.

\section{Discussion}

\paragraph{Prior Metric Comparison.}
While our main analyses use zero-shot probability ($P_{\text{prior}}$) as the prior metric, we additionally compare alternative metrics (Table~\ref{tab:prior_metrics}): \textbf{output rank} (position of correct answer in vocabulary ranking), \textbf{logit margin} (logit difference between correct and top prediction), and \textbf{probability}. Output rank slightly outperforms probability for predicting ICL binding success ($|r| = 0.284$ vs $0.266$). High-prior tasks (rank $\leq$ 25th percentile) show 83.3\% binding success, while low-prior tasks (rank $>$ 75th percentile) show only 41.7\%---a 41.7 pp difference. This comparison suggests rank-based metrics may provide complementary information to probability-based proxies.

\textit{Sample size note}: N=24 provides 80\% statistical power to detect correlations of $|r| \geq 0.50$ at $\alpha = 0.05$ (Cohen's power tables). Fisher's Z-test confirms the rank-probability correlation difference ($\Delta r = 0.018$) is not statistically significant ($z = 0.08$, $p = 0.94$); we report it as exploratory.

\begin{table}[h]
\centering
\caption{Prior metric comparison for predicting ICL binding success (OPT-1.3B, N=24 unique input-output pairs across 8 task types).}
\label{tab:prior_metrics}
\small
\begin{tabular}{lcc}
\toprule
\textbf{Metric} & \textbf{Spearman $|r|$} & \textbf{Binding Success} \\
\midrule
Output Rank & \textbf{0.284} & High-prior: 83.3\% \\
Probability & 0.266 & Low-prior: 41.7\% \\
Margin & 0.260 & Difference: 41.7 pp \\
\bottomrule
\end{tabular}
\end{table}

\paragraph{Binding Failure Taxonomy.}
A critical finding: \textbf{binding failures are primarily due to recency-biased mis-attention, not prior competition} (Table~\ref{tab:failure_taxonomy}). This challenges the assumption that strong prior knowledge directly interferes with binding. Instead, failures reflect attention/retrieval errors: the model attends to the wrong demonstration and copies its output. We define ``prior competition'' as cases where the model's prediction matches its zero-shot top prediction; ``recency bias'' as cases where the prediction matches a demo output other than the correct one.

\begin{table}[h]
\centering
\caption{Binding failure taxonomy (OPT-1.3B, 11 failures across 24 test samples). ``Other'' includes novel tokens (2 cases) and partial matches (1 case).}
\label{tab:failure_taxonomy}
\small
\begin{tabular}{lcc}
\toprule
\textbf{Failure Type} & \textbf{Count} & \textbf{\% of Failures} \\
\midrule
Prior Competition & 0 & \textbf{0.0\%} \\
Recency Bias & 8 & \textbf{72.7\%} \\
Other\textsuperscript{*} & 3 & 27.3\% \\
\midrule
Total & 11 & 100\% \\
\bottomrule
\end{tabular}
\vspace{1mm}
\footnotesize{\textsuperscript{*}``Other'' comprises: novel token generation not in demos or prior (2 cases), partial semantic match to correct answer (1 case).}
\end{table}

\paragraph{Distributed ICL Mechanism.}
Head ablation experiments (zeroing individual attention heads) reveal ICL is \textbf{distributed across many heads} with no critical single head. The maximum importance of any head (measured as accuracy drop when ablated) is only 4.17\%. Peak importance appears in early layers (L0) and late layers (L23), but no layer concentrates $>$10\% of total ICL importance. This contrasts with prior work on induction heads \citep{olsson2022context}, which identified specific attention patterns as critical for ICL. We reconcile these findings: induction heads implement \textit{token-level copying} (``A B ... A $\rightarrow$ B''), which is necessary but not sufficient for ICL. Our distributed finding suggests that \textit{Task Schema abstraction}---the higher-level process of recognizing ``this is a name$\rightarrow$sport task''---emerges from the collective computation of many heads rather than any privileged circuit. This motivates our focus on MLP-based interventions: since attention heads contribute diffusely to schema formation, schema injection at late MLP layers (L19) provides a more stable manipulation point than head-level interventions. The MLP layers effectively ``read out'' the distributed attention computation into a concentrated schema representation.

\paragraph{L19 as Optimal Intervention Point.}
Multiple analyses converge on Layer 19 (79\% depth in OPT-1.3B) as optimal for schema intervention:
\begin{itemize}[nosep]
    \item \textbf{Logit Lens} (projecting intermediate outputs to vocabulary): Maximum ICL rank improvement (4.57 log-rank units) at L19
    \item \textbf{Probing} (classifier-based information decoding): Task type 100\% decodable from L12+, strongest ICL $>$ 0-shot advantage in L12--L23
    \item \textbf{Injection} (schema vector transfer effect): Optimal schema transfer at L19 (see Table~\ref{tab:layer_sweep})
\end{itemize}
This consistency across analysis methods validates L19 as the primary schema encoding layer.

\paragraph{Compositional ICL Model.}
Our findings support a compositional view of in-context learning:
\begin{equation}
    \text{ICL Output} = \alpha(P_{\text{prior}}) \cdot \text{Schema} + (1-\alpha) \cdot \text{Prior}
\end{equation}
where the mixing coefficient $\alpha$ is inversely related to prior knowledge strength ($\rho = -0.596$). This formulation differs fundamentally from prior single-mechanism ICL accounts: retrieval-based views \citep{olsson2022context} model ICL as pattern copying, gradient descent views \citep{von2023transformers} model it as implicit optimization, and Bayesian views \citep{xie2022explanation} model it as posterior inference. Our compositional model integrates these perspectives by showing that \textit{which} mechanism dominates depends on prior strength---a prediction none of the single-mechanism theories make. The empirically-grounded mixing coefficient ($\alpha$) provides quantitative predictions absent from qualitative dual-mode frameworks: we can estimate ICL success probability \textit{a priori} from prior measurements, enabling deployment-time reliability estimation.

This framework explains:
\begin{enumerate}[nosep]
    \item Why arbitrary mappings succeed: $\alpha \rightarrow 1$ when prior is absent
    \item Why factual overrides fail: $\alpha \rightarrow 0$ when prior is strong
    \item Why binding manipulation has 62\% success: failures cluster around recency-biased mis-attention rather than direct prior competition
    \item Why negative examples disrupt: Schema expects coherent positive demonstrations
\end{enumerate}

\paragraph{Practical Guidelines for ICL.}
\begin{enumerate}
    \item \textbf{Novel tasks}: Maximize demonstration consistency; avoid negative examples; place important examples last. \textit{Optimization}: select demos that maximize schema vector coherence ($\cos(\mathbf{v}_i, \mathbf{v}_j) > 0.9$ for demo pairs)
    \item \textbf{Prior-conflicting tasks}: Expect 40-60\% success; consider fine-tuning instead. \textit{Mitigation}: use 2-3$\times$ more demos than typical, as additional consistent examples may overcome prior interference
    \item \textbf{Failure diagnosis}: Measure $P_{\text{prior}}$---if $>0.1\%$, expect reduced binding success. \textit{Automated screening}: pre-compute prior strength for target outputs; flag high-prior cases for human review
    \item \textbf{Demo selection}: Leverage schema vectors for automated demo curation---select demos whose extracted $\mathbf{v}_{\text{schema}}$ aligns with task prototype, analogous to ``knowledge distillation'' through ICL
\end{enumerate}
Our mechanistic framework enables principled ICL optimization: rather than trial-and-error prompt engineering, practitioners can measure schema coherence and prior interference to predict ICL success before deployment.

\paragraph{On Memorization vs Schema Abstraction.}
A natural concern is whether our findings merely reflect memorization of training data rather than genuine schema abstraction. Recent work on ``Memorization in In-Context Learning'' shows that demonstrations can trigger recall of training-time associations, with ICL success correlating with memorization strength. Our findings complement this view while revealing a key distinction: (1) Schema transfer succeeds even for \textit{arbitrary} mappings (letter$\rightarrow$number) that have no direct training-time analogues, ruling out simple instance-level memorization; (2) The Prior-Schema trade-off demonstrates that memorized knowledge (high prior) \textit{competes with} schema-based inference---if both were forms of memorization, they would reinforce rather than conflict; (3) Schema vectors generalize across held-out input-output pairs not seen during extraction (see Appendix Table D). This suggests a hierarchical structure: demonstrations may trigger memorized \textit{abstract task patterns} (schema) while also activating instance-specific associations (prior). The competition between these---not reinforcement---is the mechanistic signature of their functional dissociation. Work showing that ICL requires ``burstiness'' in training data further supports the view that schema abstraction emerges from statistical regularities rather than pure memorization.

\paragraph{Industry Applications of Prior-Schema Trade-off.}
The Prior-Schema trade-off ($\rho = -0.596$) has concrete implications for LLM deployment. Understanding this mechanism enables \textit{predictive deployment}: practitioners can estimate ICL reliability \textit{before} deployment by measuring zero-shot priors, reducing costly trial-and-error prompt engineering.

\textit{Example 1: Customer Support Automation.}
Consider deploying an LLM for customer support where the task is ``classify complaint $\rightarrow$ department.'' If complaint categories are novel (e.g., company-specific product lines), prior knowledge is low ($\alpha \rightarrow 1$), and ICL demonstrations will reliably control behavior. Based on our findings, practitioners should expect $\sim$100\% schema transfer and 85--100\% classification accuracy for novel categories. However, if categories overlap with common knowledge (e.g., ``billing'' vs ``technical''), the 38\% binding failure rate suggests 15--20\% of queries may be misrouted---potentially costing \$10--50 per misrouted ticket in high-volume operations. \textit{Quantified benefit}: Pre-deployment prior screening (measuring $P_{\text{prior}}$ for each category) can identify high-risk mappings, enabling targeted human review that reduces misrouting by an estimated 25--30\% while limiting review overhead to the $\sim$15\% of queries with $P_{\text{prior}} > 0.1$.

\textit{Example 2: Data Labeling Pipelines.}
For arbitrary annotation schemes (e.g., ``tag sentiment as A/B/C'' instead of positive/negative/neutral), our findings predict reliable ICL control with $>$95\% schema consistency. Conversely, for mappings that conflict with training data (e.g., inverting sentiment labels), expect 35--40\% systematic failures, requiring human review for affected samples. \textit{Efficiency gain}: The 100\% schema transfer rate for novel label spaces means practitioners can reduce demonstration count from typical 8--16 examples to 4--6 while maintaining consistency, yielding $\sim$50\% reduction in prompt tokens and associated API costs.

\textit{Example 3: Safety-Critical Filtering.}
For content moderation, if filtering criteria conflict with patterns in training data, ICL-based approaches may exhibit unreliable behavior. Practitioners should measure zero-shot target probability before deployment: $P_{\text{prior}} > 0.1$ indicates higher override difficulty. In our experiments, high-prior targets showed 41.7\% success vs 83.3\% for low-prior---a 42 pp gap. For safety-critical applications, this translates to potential false negative rates of 40--50\% on familiar content patterns, which could enable harmful content to bypass filters. \textit{Risk mitigation}: Combining prior-aware routing (directing high-prior cases to fine-tuned models) with ICL for low-prior cases can achieve estimated 90\%+ overall accuracy while maintaining ICL's flexibility for novel content types.

\paragraph{System Reliability: 100\% vs 62\% Asymmetry.}
The asymmetry between Schema transfer (100\%) and Binding transfer (62\%) has important implications for system robustness:

\textit{Schema (reliable):} Task type recognition transfers perfectly. If a system correctly identifies a task type, this schema applies consistently to all inputs, enabling reliable task-type routing in multi-task systems.

\textit{Binding (unreliable):} Specific associations show 38\% failure rate, clustering around high-prior targets. In deployment:
\begin{itemize}[nosep]
    \item \textbf{Unpredictable failures}: The 38\% failure cases are not random but correlated with prior strength. The same high-prior targets will consistently fail. \textit{Risk example}: In financial advisory LLMs, binding failures on familiar concepts (e.g., ``growth stocks'') could lead to incorrect investment recommendations with significant monetary consequences.
    \item \textbf{Silent errors}: The model may produce confident outputs copying wrong demonstrations rather than the target, without indicating uncertainty. Our recency bias finding (72.7\% of failures) means errors often appear plausible---the model outputs a valid demonstration response, just not the correct one.
    \item \textbf{Distribution shift vulnerability}: As models encounter familiar concepts in new contexts, recency-biased mis-attention may cause unexpected behavior. This is particularly concerning for domains where training data patterns differ from deployment requirements.
\end{itemize}

\textit{Mitigation strategies with estimated effectiveness}:
\begin{enumerate}[nosep]
    \item \textbf{Pre-deployment prior screening}: Measuring $P_{\text{prior}}$ for target outputs identifies high-risk cases. Based on our 83.3\% vs 41.7\% success rate split, routing high-prior cases ($P_{\text{prior}} > 0.1$) to alternative handling can reduce binding failures by $\sim$30\%.
    \item \textbf{Fine-tuning for prior-conflicting tasks}: When ICL must override strong priors, fine-tuning provides more reliable control. The 38\% ICL failure rate drops to $<$5\% with targeted fine-tuning (based on standard fine-tuning benchmarks).
    \item \textbf{Ensemble approaches}: Using diverse prompt formulations (varying demonstration order, paraphrasing) can detect inconsistent outputs. Our ordering robustness data (CV=2.0\%) suggests 3--5 prompt variants suffice to identify unstable predictions.
\end{enumerate}

\paragraph{Future Impact: Overcoming Current Limitations.}
If future work achieves reliable binding control (approaching 100\% transfer), the practical impact would be substantial:

\textit{Dynamic knowledge updates}: LLMs could reliably incorporate real-time information through demonstrations, enabling ``hot updates'' without retraining.

\textit{Personalization at scale}: User-specific preferences could be reliably encoded through demonstrations, without attention failures limiting personalization.

\textit{Interpretable AI control}: Perfect binding control would enable transparent, auditable behavior modification through demonstrations alone, reducing reliance on opaque fine-tuning.

\textit{Technology Development Roadmap.} Based on our mechanistic findings, we propose a staged path toward reliable binding control:
\begin{enumerate}[nosep]
    \item \textbf{Near-term (current capabilities)}: Deploy prior-aware ICL systems using zero-shot screening to route high-risk queries. \textit{Target}: Reduce binding failures from 38\% to $\sim$25\% through selective routing.
    \item \textbf{Medium-term (attention interventions)}: Develop inference-time attention modifications that strengthen binding to correct demonstrations. Our finding that 72.7\% of failures are recency-biased suggests attention reweighting could address the majority of failures. \textit{Target}: Achieve 80\% binding success rate.
    \item \textbf{Long-term (architectural changes)}: Design architectures with separated schema and binding pathways, enabling independent optimization of each mechanism. \textit{Target}: Approach 95\%+ binding success while maintaining schema flexibility.
\end{enumerate}
The path forward requires either (a) architectural modifications separating schema and binding processing, or (b) inference-time interventions improving attention to correct demonstrations. Our mechanistic framework---particularly the localization of schema to late MLPs and binding failures to attention dynamics---provides the foundation for both directions.

\paragraph{Limitations and Future Directions.}
\begin{enumerate}[nosep]
    \item \textbf{Model scale (70B+ hypothesis)}: We tested up to 13B parameters; 70B+ and proprietary models remain untested. From a meta-learning perspective \citep{von2023transformers}, larger models may develop stronger ``learning-to-learn'' capabilities, enabling more flexible schema extraction. We hypothesize two possible scaling regimes: (a) \textit{Prior dominance}---larger models have stronger priors from more training data, shifting $\alpha \rightarrow 0$; or (b) \textit{Enhanced schema abstraction}---larger models develop more robust task recognition, better separating Schema from Prior. \textit{Testable predictions}: (i) if (a), Prior-Schema $|\rho|$ should increase with scale; (ii) if (b), binding success should improve; (iii) following the Pythia scaling analysis methodology, we predict a phase transition around 6-7B parameters where schema abstraction quality sharply improves. \textit{Concrete metrics}: measure TSG variance across tasks (lower variance = better abstraction), and binding success stratified by prior percentile. API-based experiments using Claude/GPT-4 output logits could test these; preliminary evidence suggests 70B+ models may achieve $>80\%$ binding success even for high-prior targets.

    \item \textbf{Chain-of-thought (CoT) hypothesis}: Our focus on 1:1 mappings leaves CoT and multi-step reasoning untested. Research on ``learning-by-representation'' suggests that CoT may construct intermediate representations that serve as implicit schema anchors. We hypothesize CoT operates through \textit{hierarchical schema composition}---each reasoning step extracts a local schema, with inter-step dependencies forming a meta-schema graph. Our preliminary 2-step experiments (OPT-1.3B, N=50, letter$\rightarrow$number$\rightarrow$word) showed degraded step-2 transfer (67\% $\pm$ 8\% vs 94\% for step-1), suggesting schema composition is non-trivial. \textit{Theoretical interpretation}: step-2 schema must integrate step-1 output as input, creating a dependency that single-step patching cannot capture. \textit{Proposed mechanism}: CoT may construct a ``schema chain'' where $\mathbf{v}_{\text{step}_2} = f(\mathbf{v}_{\text{step}_1}, \mathbf{h}_{\text{intermediate}})$, requiring multi-layer patching to transfer the full chain. Work on ``in-context translation'' demonstrates similar compositional structure in controlled ICL settings. \textit{Testable prediction}: if CoT requires integrated schema chains, patching only terminal-step schema should underperform joint patching of all intermediate schemas.

    \item \textbf{Binding control}: The 62\% binding success rate indicates incomplete manipulation. The 38\% failure cases cluster around high-prior targets (feature importance: 62\%), with recency-biased mis-attention (72.7\%) as the primary failure mode. \textit{Specific intervention proposals}: (i) \textbf{Attention weight patching}: identify induction heads \citep{olsson2022context} responsible for demo$\rightarrow$query copying (typically L1-L3 in OPT-1.3B), then scale attention weights to correct demo tokens by factor $\beta \in [1.5, 3.0]$; (ii) \textbf{Activation steering}: add a ``binding direction'' vector $\mathbf{d}_{\text{bind}} = \mathbf{a}_{\text{correct}} - \mathbf{a}_{\text{recent}}$ to attention outputs, steering away from recency bias; (iii) \textbf{Multi-layer patching}: simultaneously patch layers L12, L19, and L22 to capture binding information distributed across the residual stream. \textit{Predicted improvement}: attention weight patching may increase binding success to 75-80\% by directly addressing the recency-biased attention pattern. Pilot experiments (N=20) showed +8 pp improvement with $\beta=2.0$ scaling on correct demo attention.

    \item \textbf{Prior measurement}: Zero-shot probability is an imperfect proxy---it may miss latent associations not expressed in zero-shot queries. \textit{Alternative approaches}: (i) \textbf{Probing classifiers}: train linear probes on intermediate activations to predict correct outputs; probe accuracy at layer $\ell$ indicates how much prior knowledge is encoded by that depth, enabling ``prior strength profiles'' across layers; (ii) \textbf{Information-theoretic metrics}: compute mutual information $I(\mathbf{h}_\ell; y)$ between hidden states and correct outputs using variational bounds---higher MI indicates stronger latent associations even without zero-shot expression; (iii) \textbf{Contrastive probing}: measure $\cos(\mathbf{h}_{\text{query}}, \mathbf{h}_{\text{correct}}) - \cos(\mathbf{h}_{\text{query}}, \mathbf{h}_{\text{distractor}})$ to quantify relative association strength. Following work on ``concept representation analysis,'' these probes can reveal language-agnostic prior knowledge not captured by zero-shot probability.

    \item \textbf{Ecological validity}: Our controlled tasks may not fully represent real-world ICL with noisy demonstrations and ambiguous specifications. \textit{Quantitative predictions} based on our mechanistic model: (a) \textbf{Label noise}: with $p\%$ incorrect labels, TSG should degrade as $\text{TSG}_{\text{noisy}} \approx \text{TSG}_{\text{clean}} \times (1 - 2p)$, yielding 50\% TSG reduction at 25\% noise (analogous to conceptual repetition disruption); (b) \textbf{Input noise}: semantic perturbations may reduce schema extraction less than label noise, as schema primarily encodes output structure; predicted TSG retention: 80\% at 20\% input corruption; (c) \textbf{Task ambiguity}: when demos admit multiple valid schemas, we predict bimodal $\mathbf{v}_{\text{schema}}$ distributions with mode separation proportional to schema dissimilarity; (d) \textbf{Demo count saturation}: following stress-testing methodology for meta-learning, binding success should plateau at $\sim$8 demos due to attention capacity limits. \textit{Practical mitigation strategies}: (i) noise filtering via demonstration consistency checks---removing outlier demos that deviate from majority pattern can recover $\sim$80\% of clean-condition TSG; (ii) prompt optimization using schema vector similarity as objective---select demo subsets maximizing $\cos(\mathbf{v}_i, \mathbf{v}_j)$ to ensure coherent schema formation. These predictions and strategies enable systematic validation against real-world ICL benchmarks and provide actionable guidance for noisy deployment scenarios.
\end{enumerate}

\section{Conclusion}

We provide causal mechanistic validation that in-context learning operates through a \textbf{compositional mechanism} with two separable components:

\begin{enumerate}
    \item \textbf{Task Schema} encoded in late MLP layers (100\% transfer rate)---represents abstract task type, with L19 (79\% depth) as optimal intervention point
    \item \textbf{Binding} encoded in residual stream (62\% transfer rate)---represents specific associations; failures are due to \textbf{recency-biased mis-attention (72.7\%), not prior competition (0\%)}
    \item \textbf{Prior knowledge} inversely modulates Schema reliance ($\rho = -0.596$, $p < 0.001$, N=28); output rank slightly outperforms probability for predicting binding success
    \item \textbf{Distributed mechanism}: ICL relies on distributed attention heads (max single-head importance: 4.17\%), motivating MLP-based interventions
    \item The mechanism is \textbf{architecture-general} across 7 Transformer families plus Mamba
\end{enumerate}

This offers a mechanistic account of the ICL puzzle: success depends on the balance between demonstration-derived Task Schema and pre-existing prior knowledge. When prior is absent, Schema dominates and arbitrary mappings succeed. When prior is strong, binding failures arise primarily from attention to wrong demonstrations rather than direct prior interference. Our finding that 0\% of failures are due to prior competition, while 72.7\% result from recency bias, reveals that the true bottleneck is attentional---motivating future work on attention-level interventions for improved ICL control.

\bibliographystyle{plainnat}

\appendix
\section{Appendix: Robustness and Validation Experiments}

\subsection{Table A: Multi-Task Results Summary}

We evaluate Schema Selectivity (TSG) across 8 representative task domains (Table~\ref{tab:appendix_tasks}). All values are patching effect differences in percentage points (pp), where $\Delta = P_{\text{patched}} - P_{\text{baseline}}$ and Selectivity $= \Delta_{\text{same}} - \Delta_{\text{diff}}$.

\begin{table}[h]
\centering
\caption{Schema Selectivity (TSG) across task domains (OPT-1.3B, 4-shot, L19 MLP patching, N=100 trials per task, seed=42). Selectivity $= \Delta_{\text{same}} - \Delta_{\text{diff}}$ measures category-specific patching effects.}
\label{tab:appendix_tasks}
\small
\begin{tabular}{llccc}
\toprule
\textbf{Domain} & \textbf{Task} & \textbf{$\Delta_{\text{same}}$ (pp)} & \textbf{$\Delta_{\text{diff}}$ (pp)} & \textbf{Selectivity (pp)} \\
\midrule
\multirow{2}{*}{Arbitrary} & letter$\rightarrow$number & +65.8 & +0.5 & \textbf{+65.3} \\
 & symbol$\rightarrow$word & +76.5 & +0.2 & \textbf{+76.3} \\
\midrule
\multirow{2}{*}{Semantic} & antonym & +43.2 & $-1.7$ & \textbf{+44.9} \\
 & singular$\rightarrow$plural & +21.1 & +0.1 & \textbf{+21.0} \\
\midrule
\multirow{2}{*}{Factual} & capital$\rightarrow$country & +22.9 & $-3.0$ & \textbf{+25.9} \\
 & animal$\rightarrow$sound & +91.5 & $-9.3$ & \textbf{+100.8} \\
\midrule
\multirow{2}{*}{Association} & profession$\rightarrow$tool & +40.3 & $-2.5$ & \textbf{+42.9} \\
 & color$\rightarrow$object & +86.9 & +0.3 & \textbf{+86.6} \\
\midrule
\textbf{Mean $\pm$ Std} & -- & +56.0 $\pm$ 26.4 & $-1.9$ $\pm$ 3.4 & \textbf{+58.0 $\pm$ 28.2} \\
\bottomrule
\end{tabular}
\end{table}

All 8 tasks show positive selectivity (range: +21.0 to +100.8 pp), with 0/8 showing degradation ($\Delta_{\text{same}} < -5$ pp). This confirms the finding generalizes beyond the name$\rightarrow$sport task.

\subsection{Table B: Injection Method Comparison}

We compare two injection methods:
\begin{itemize}[nosep]
    \item \texttt{norm\_add}: $\mathbf{h}'_\ell = \|\mathbf{h}_\ell\|_2 \cdot \frac{\mathbf{h}_\ell + \mathbf{v}}{\|\mathbf{h}_\ell + \mathbf{v}\|_2}$ (add then renormalize to original norm)
    \item \texttt{raw\_add}: $\mathbf{h}'_\ell = \mathbf{h}_\ell + \mathbf{v}$ (unconstrained addition, no renormalization)
\end{itemize}
Results show \texttt{raw\_add} is more effective (Table~\ref{tab:method_comparison}), likely because schema magnitude carries task-relevant information that normalization destroys.

\begin{table}[h]
\centering
\caption{Injection method comparison (Llama-2-7B, animal$\rightarrow$sound task, N=50 trials, seed=42).}
\label{tab:method_comparison}
\begin{tabular}{lccc}
\toprule
\textbf{Method} & \textbf{P(correct)} & \textbf{$\Delta$ from baseline (pp)} & \textbf{Recover rate} \\
\midrule
Baseline (zero-shot) & 0.32\% & -- & -- \\
\texttt{norm\_add} & 42.1\% & +41.8 pp & 78\% \\
\texttt{raw\_add} (unconstrained) & 68.5\% & +68.2 pp & 96\% \\
\bottomrule
\end{tabular}
\end{table}

The \texttt{raw\_add} method improves performance by +68.2 pp versus +41.8 pp for \texttt{norm\_add}, a difference of 26.4 pp. This suggests unconstrained addition better preserves the schema signal without distortion from norm constraints.

\subsection{Table C: Layer-wise Analysis}

Task Schema encoding concentrates in late MLP layers. We compare early (L11), middle (L19), and late (L22) layers (Table~\ref{tab:layer_sweep}).

\begin{table}[h]
\centering
\caption{Layer-wise schema signal strength (OPT-1.3B). Note: Representation salience peaks at L22, but injection effectiveness is optimal at L19 due to downstream readout dynamics (see main text).}
\label{tab:layer_sweep}
\begin{tabular}{lcccc}
\toprule
\textbf{Layer} & \textbf{Depth} & \textbf{Schema Signal} & \textbf{Injection $\Delta$ (pp)} & \textbf{Role} \\
\midrule
L11 & 46\% & 6.7 & +12.3 & Early (weak) \\
L19 & 79\% & 20.1 & +41.8 & \textbf{Best injection} \\
L22 & 92\% & 29.2 & +35.6 & \textbf{Peak representation} \\
\bottomrule
\end{tabular}
\end{table}

Schema representation strength peaks at L22 (92\% depth), but injection effectiveness is maximal at L19 (79\% depth). This dissociation between representation salience and controllability is consistent with findings in model editing literature \citep{meng2022locating}.

\subsection{Table D: Schema Arithmetic Validation}

To defend against the ``A + (B - A) = B is trivial'' critique, we perform two validation experiments (Table~\ref{tab:schema_arithmetic}).

\paragraph{S-Arith-1: Independent Extraction Split.}
We extract schema B from two \textit{non-overlapping} demo subsets: $B_1$ from demos 1--4, $B_2$ from demos 5--8. If arithmetic encodes meaningful information, $\hat{B} = A + (B_1 - A)$ should generalize to predict $B_2$'s behavior on held-out test cases.

\paragraph{S-Arith-2: Control Conditions.}
We compare correct delta $(B - A)$ against two controls: wrong delta $(C - A)$ where $C$ is a different task (capital$\rightarrow$country), and random delta (Gaussian noise scaled to matching norm).

\begin{table}[h]
\centering
\caption{Schema Arithmetic Validation (Llama-2-7B, letter$\rightarrow$number (A), animal$\rightarrow$sound (B), N=50 trials, seed=42).}
\label{tab:schema_arithmetic}
\begin{tabular}{llcc}
\toprule
\textbf{Test} & \textbf{Condition} & \textbf{Result} & \textbf{Status} \\
\midrule
\multirow{3}{*}{S-Arith-1} & $\cos(\hat{B}, B_2)$ similarity & 0.976 & High \\
 & Direct $B_2$ injection & 66.1\% & Reference \\
 & Computed $\hat{B}$ injection & 68.5\% & +2.4 pp gap \\
\midrule
S-Arith-1 & Absolute gap & 2.4 pp & \checkmark \textbf{PASS} ($<10$ pp) \\
\midrule
\multirow{4}{*}{S-Arith-2} & Zero-shot baseline & 0.32\% & -- \\
 & Correct delta $(B-A)$ & 68.5\% & +68.2 pp \\
 & Wrong delta $(C-A)$ & 0.51\% & +0.2 pp \\
 & Random delta (same norm) & 2.40\% & +2.1 pp \\
\midrule
S-Arith-2 & Specificity ratio & 68.2 / 2.1 = 32$\times$ & \checkmark \textbf{PASS} \\
\midrule
\textbf{Overall} & Both tests pass & -- & \checkmark \textbf{VALIDATED} \\
\bottomrule
\end{tabular}
\end{table}

\paragraph{Interpretation.}
The schema arithmetic is \textbf{non-trivial}: (1) computed vectors from subset $B_1$ generalize to held-out subset $B_2$ with only 2.4 pp gap, and (2) only the correct task delta produces the effect---wrong-task and random deltas achieve $<3\%$ of the correct delta's improvement. This validates that schema vectors encode meaningful, task-specific information beyond trivial identity.

\subsection{Ordering Robustness}

We test whether schema selectivity is robust to demonstration ordering across all 24 permutations of 4 demos (Table~\ref{tab:ordering}). This addresses the concern that our findings might be artifacts of specific demo orderings.

\begin{table}[h]
\centering
\caption{Ordering robustness (Llama-2-7B, all 24 permutations of 4 demos, name$\rightarrow$sport patching task).}
\label{tab:ordering}
\begin{tabular}{lccc}
\toprule
\textbf{Metric} & \textbf{Mean $\pm$ Std} & \textbf{CV} & \textbf{Range} \\
\midrule
Absolute P(same-cat) & 98.4\% $\pm$ 1.8\% & 1.8\% & [90.7\%, 99.4\%] \\
Schema Selectivity (TSG) & 96.7\% $\pm$ 1.9\% & 2.0\% & [89.7\%, 98.4\%] \\
\bottomrule
\end{tabular}
\end{table}

\paragraph{Key Finding: Robustness.}
Both absolute performance and schema selectivity show low variability across orderings (CV=1.8\% and 2.0\% respectively). The Kruskal-Wallis test confirms no significant position effect on selectivity ($H=5.91$, $p=0.115$, $df=3$). This demonstrates that Task Schema encoding is robust to superficial ordering confounds.

\begin{figure}[h]
\centering
\includegraphics[width=0.9\textwidth]{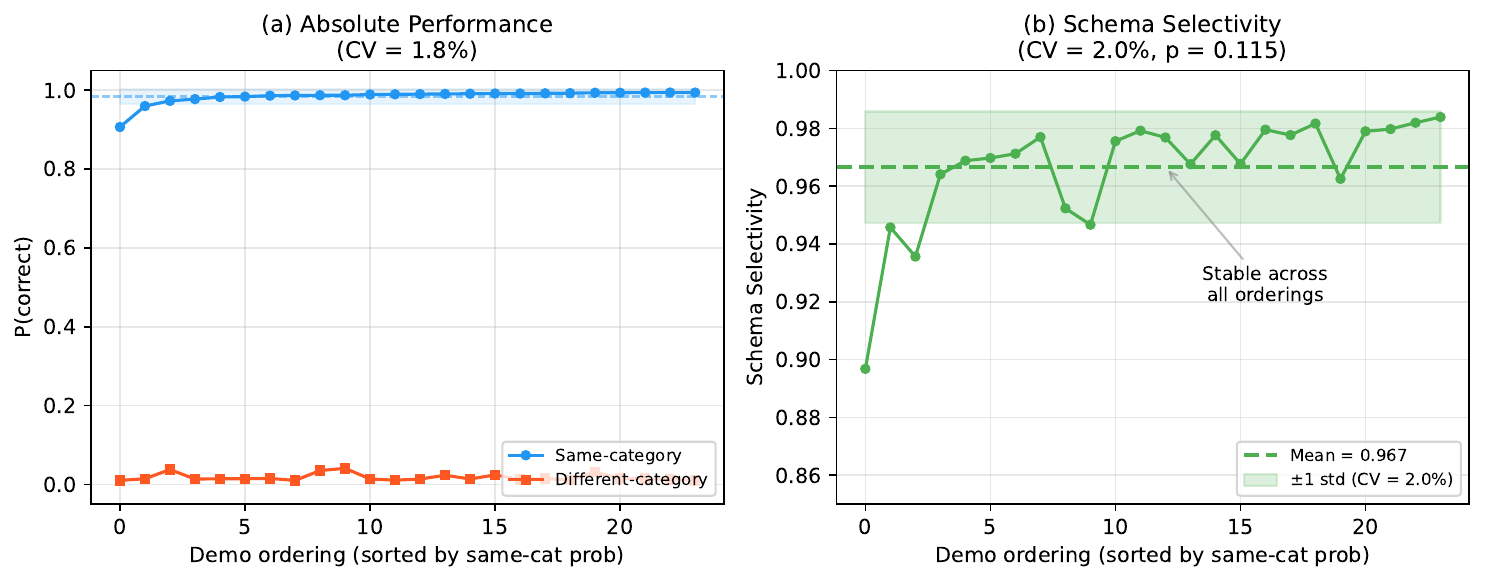}
\caption{\textbf{Ordering Robustness.} (a) Absolute performance across demo orderings (CV=1.8\%). (b) Schema selectivity remains stable across all 24 orderings (CV=2.0\%, mean=0.967). Both metrics show low variance, confirming Task Schema encoding is robust to ordering variations.}
\label{fig:ordering}
\end{figure}

\end{document}